\documentclass{llncs}
\usepackage{amsmath}
\usepackage{bbm}
\usepackage{epsfig}
\usepackage{psfrag}

\newcommand{\F}{\ensuremath{\mathcal F}}
\renewcommand{\H}{\ensuremath{\mathcal H}}
\newcommand{\pot}{\ensuremath{\Pi}}
\newcommand{\X}{\ensuremath{\mathcal X}}
\newcommand{\Y}{\ensuremath{\mathcal Y}}
\newcommand{\m}{d}
\newcommand{\clup}{c.u.p.}

\newcommand{\geno}{\X}
\newcommand{\x}{|\geno|} 
\newcommand{\f}{f} 
\newcommand{\opt}{{n}}
 
\newcommand{\ti}{{\mathcal T}_{\x,\opt}} 

\newcommand{\dN}{\mathbbm N}
\newcommand{\dR}{\mathbbm R}

\begin{document}
\title{\textbf{Recent Results on\\No-Free-Lunch Theorems for Optimization}}
\author{Christian Igel \and Marc Toussaint}
\institute{Institut f\"ur Neuroinformatik\\Chair of Theoretical Biology\\Ruhr-Universit\"at Bochum\\44780 Bochum, Germany\\\texttt{\{igel,toussaint\}@neuroinformatik.rub.de}}
\date{}
\maketitle

\hyphenation{ex-plo-ra-tion}

\renewcommand{\theenumi}{\alph{enumi}}
\renewcommand{\labelenumi}{(\alph{enumi})}

\begin{abstract}\normalsize
\noindent 
The sharpened No-Free-Lunch-theorem (NFL-theorem) states that the
performance of all optimization algorithms averaged over any finite
set $F$ of functions is equal if and only if $F$ is closed under
permutation (\clup) and each target function in $F$ is equally likely.
In this paper, we first summarize some consequences of this theorem,
which have been proven recently: The average number of evaluations
needed to find a desirable (e.g., optimal) solution can be calculated;
the number of subsets \clup\ can be neglected compared to the overall
number of possible subsets; and problem classes relevant in practice
are not likely to be \clup{} Second, as the main result, the
NFL-theorem is extended.  Necessary and sufficient conditions for
NFL-results to hold are given for arbitrary, non-uniform distributions
of target functions.  This yields the most general NFL-theorem for
optimization presented so far.

\end{abstract}

\section{Introduction}
Search heuristics such as evolutionary algorithms, grid search,
simulated annealing, and tabu search are general in the sense that they
can be applied to any target function $f:\X\to\Y$, where $\X$ denotes
a finite search space and $\Y$ is a finite set of totally ordered
cost-values. Much research is spent on developing search heuristics
that are superior to others when the target functions belong to a
certain class of problems. But under which conditions can one search
method be better than another?  The No-Free-Lunch-theorem for
optimization (NFL-theorem) roughly speaking states that all
non-repeating search algorithms have the same mean performance when
averaged uniformly over \emph{all} possible objective functions
$f:\X\to\Y$
\cite{wolpert:95,radcliffe:95,wolpert:97,koeppen:01,droste:02}.  Of
course, in practice an algorithm need not perform well on all possible
functions, but only on a subset that arises from the real-world
problems at hand, e.g., optimization of neural networks. Recently, a
sharpened version of the NFL-theorem has been proven that states that
NFL-results hold (i.e., the mean performance of all search algorithms
is equal) for any subset $F$ of the set of all possible functions if
and only if $F$ is closed under permutation (\clup) and each target
function in $F$ is equally likely \cite{schumacher:01}.

In this paper, we address the following basic questions:
When all algorithms have the same mean performance---how long
does it take on average to find a desirable solution?
How likely is it that a randomly chosen subset of functions
is \clup, i.e., fulfills the prerequisites
of the sharpened NFL-theorem?
Do constraints relevant in practice lead to classes of target functions
that are \clup?
And finally: How can the NFL-theorem be extended to non-uniform
distributions of target functions?  Answers to all these questions are
given in the sections \ref{timetodesire:sec} to \ref{toppology:sec}.
First, the scenario considered in NFL-theorems is described formally.

\section{Preliminaries}\label{prem:sec}

\begin{figure}
\small
\psfrag{a}[][]{$c(Y(f,m,a))$}
\psfrag{s}[][]{performance measure}
\psfrag{b}[][]{ non-repeating black-box search algorithm $a$}
\psfrag{c}[][]{$T_m=\left<(x_1,f(x_1)),\dots, (x_m,f(x_m))\right>$}
\psfrag{d}[][]{{target function}}
\psfrag{u}[][]{{$f:\X\to\Y$}}
\psfrag{e}[r][r]{\small $Y(f,m,a)=$}
\psfrag{t}[r][r]{$\left<f(x_1),\dots,f(x_m)\right>$}
\psfrag{f}[r][r]{$x_{m+1}\in\X$}
\psfrag{g}[l][l]{$f(x_{m+1})\in\Y$}
\centerline{\epsfig{file=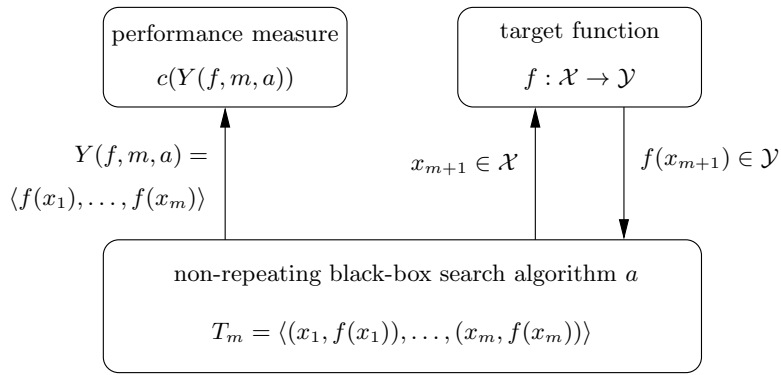,width=.66\textwidth}}

\caption{\label{blackbox:fig}Schema of the optimization scenario considered
  in NFL-theorems. A non-repeating black-box search algorithm $a$
  chooses a new exploration point in the search space depending on the
  sequence $T_m$ of the already visited points with their
  corresponding cost-values.  The target function $f$ returns the
  cost-value of a candidate solution as the only information.  The
  performance of $a$ is determined using the performance measure $c$,
  which is a function of the sequence $Y(f,m,a)$ containing the cost-values.}
\end{figure}

A finite search space $\X$ and a finite set of cost-values $\Y$ are
presumed.  Let $\F$ be the set of all objective functions $f:\X\to\Y$
to be optimized (also called target, fitness, energy, or cost functions).
NFL-theorems are concerned with non-repeating black-box search
algorithms (referred to as algorithms) that choose
a new exploration point in the search space depending on the 
history of prior explorations: The sequence
$T_m=\left<(x_1,f(x_1)),(x_2,f(x_2)), \dots, (x_m,f(x_m))\right>$
represents $m$ pairs of different search points $x_i \in \X$,
$\forall  {i,j}:\, x_i\not=x_j$ and their cost-values $f(x_i) \in \Y$.
An algorithm $a$ appends a pair $(x_{m+1},f(x_{m+1}))$ to this
sequence by mapping $T_m$ to a new point $x_{m+1}$, $\forall {i}:\,
x_{m+1}\not=x_i$. In many search heuristics, such as evolutionary
algorithms or simulated annealing in their canonical form, it is not
ensured that a point in the search space is evaluated only once.
However, these algorithms can become non-repeating when
they are coupled with a search-point database, see \cite{igel:02e} for
an example in the field of structure optimization of neural networks.

The performance of an algorithm $a$ after $m$ iterations with respect
to a function $f$ depends only on the sequence
$Y(f,m,a)=\left<f(x_1),f(x_2), \dots,f(x_m)\right>$ of cost-values,
the algorithm has produced.  Let the function $c$ denote a performance
measure mapping sequences of cost-values to the real numbers. For
example, in the case of function minimization a performance measure
that returns the minimum cost-value in the sequence could be a
reasonable choice. See Fig.~\ref{blackbox:fig} for a schema of the
scenario assumed in NFL-theorems.

Using these definitions, the original NFL-theorem for optimization
reads:
\begin{theorem}[NFL-theorem \protect\cite{wolpert:97}]\label{nflorig:thm}
For any two  algorithms $a$ and $b$,
any  $k\in \dR$, any
  $m\in\{1,\dots,|\X|\}$, and any performance measure $c$
\begin{equation}\label{wolpi:eq}
\sum_{f\in \F}\delta(k, c(Y(f,m,a)))
=
\sum_{f\in \F}\delta(k, c(Y(f,m,b)))\enspace.
\end{equation}
\end{theorem}
\noindent Herein, $\delta$ denotes the Kronecker function ($\delta(i,j)=1$ if
$i=j$, $\delta(i,j)=0$ otherwise).  Proofs can be found in
\cite{wolpert:95,wolpert:97,koeppen:01}. This theorem implies that for
any two (deterministic or stochastic, cf.~\cite{droste:02}) algorithms
$a$ and $b$ and any function $f_a\in \F$, there is a function $f_b\in
F$ on which $b$ has the same performance as $a$ on $f_a$. Hence,
statements like ``Averaged over all functions, my search algorithm is
the best'' are misconceptions. Note that the summation in
(\ref{wolpi:eq}) corresponds to uniformly averaging over all functions
in $\F$, i.e., each function has the same probability to be the target
function.

Recently, theorem \ref{nflorig:thm} has been extended to subsets of
functions that are closed under permutation (\clup).  Let
$\pi:\X\rightarrow\X$ be a permutation of $\X$.  The set of all
permutations of $\X$ is denoted by $\pot(\X)$.  A set $F\subseteq\F$
is said to be \clup\ if for any $\pi\in\pot(\X)$ and any function
$f\in F$ the function $f\circ\pi$ is also in $F$.

\begin{example}
Consider the mappings $\{0,1\}^2\to\{0,1\}$, denoted by
$\f_0,\f_1,\dots,\f_{15}$ as shown in table \ref{f:tab}.  
Then the set
$\{\f_1,\f_2,\f_4,\f_8\}$ is \clup, also $\{\f_0,
\f_1,\f_2,\f_4,\f_8\}$. The set $\{\f_1,\f_2,\f_3,\f_4,\f_8\}$
is not \clup, because some functions are ``missing'', e.g.,
$\f_5$, which results from $\f_3$  by switching
the elements $(0, 1)^T$ and  $(1, 0)^T$.
\end{example}

\begin{table}[hb]
\begin{center}
\caption{\label{f:tab}Functions $\{0,1\}^2\to\{0,1\}$.}\medskip
\small
\begin{tabular}{@{}lc@{\,\,}c@{\,\,}c@{\,\,}c@{\,\,}c@{\,\,}c@{\,\,}c@{\,\,}c@{\,\,}c@{\,\,}c@{\,\,}c@{\,\,}c@
{\,\,}c@{\,\,}c@{\,\,}c@{\,\,}c@{}}
\hline
$(x_1, x_2)^T$\rule[-7pt]{0pt}{20pt} & $\f_{0\phantom{0}}$ & $\f_{1\phantom{0}}$ & $\f_{2\phantom{0}}$ &$\f_{3
\phantom{0}}$ &$\f_{4\phantom{0}}$ & $\f_{5\phantom{0}}$ & $\f_{6\phantom{0}}$
& $\f_{7\phantom{0}}$ & $\f_{8\phantom{0}}$ &$\f_{9\phantom{0}}$ &$\f_{10}$ &$\f_{11}$ &$\f_{12}$ &$\f_{13}$ &
$\f_{14}$ & $\f_{15}$ \\
\hline
$(0, 0)^T$\rule[0pt]{0pt}{13pt}& 0& 1& 0& 1 & 0& 1 & 0& 1& 0& 1& 0& 1 & 0& 1 & 0& 1\\
$(0, 1)^T$& 0& 0& 1& 1& 0& 0& 1& 1& 0& 0& 1& 1& 0& 0& 1& 1\\
$(1, 0)^T$& 0& 0& 0& 0& 1& 1 & 1 & 1&  0& 0& 0& 0& 1& 1 & 1 & 1\\
$(1, 1)^T$\rule[-7pt]{0pt}{0pt}& 0& 0& 0& 0& 0& 0& 0& 0& 1& 1 & 1 & 1& 1& 1 & 1 & 1\\
\hline
\end{tabular}
\end{center}
\end{table}

In \cite{schumacher:01} it is proven:

\begin{theorem}[sharpened NFL-theorem \protect\cite{schumacher:01}]\label{nfl:thm}
  For any two algorithms $a$ and $b$, any $k\in \dR$, any
  $m\in\{1,\dots,|\X|\}$, and any performance measure $c$
\begin{equation}
\sum_{f\in F}\delta(k, c(Y(f,m,a)))
=
\sum_{f\in F}\delta(k, c(Y(f,m,b)))
\end{equation}
iff $F$ is \clup{}
\end{theorem}
This is an important extension of theorem \ref{nflorig:thm}, 
because it gives necessary and sufficient conditions
for NFL-results for subsets of functions. But still
theorem \ref{nfl:thm} can only be applied if all elements in $\F$ have the
same probability to be the target function, because the summations average
uniformly over $F$.

In the following, the concept of $\Y$-histograms is useful.
A \emph{\Y-histogram} (\emph{histogram} for short) is a
mapping $h:\, \Y \to {\dN}_0$ such that $\sum_{y \in \Y} h(y) = |\X|$.
The set of all histograms is denoted $\H$. Any function $f:\, \X
\to \Y$ implies the histogram $h_f(y)=| f^{-1}(y)|$ that counts the
number of elements in $\X$ that are mapped to the same value $y \in
\Y$ by $f$.  Herein, $f^{-1}(y), y\in\Y$ returns the preimage
$\{x|f(x) = y\}$ of $y$ under $f$.  Further, two functions $f,g$ are called
\emph{$h$-equivalent} iff they have the same histogram.  The
corresponding $h$-equivalence class $B_h \subseteq\F$ containing all
functions with histogram $h$ is termed a \emph{basis class}.

\begin{example}
  Consider the functions in table \ref{f:tab}.  The $\Y$-histogram of
  $\f_1$ contains the value zero three times and the value one one
  time, i.e., we have $h_{\f_1}(0)=3$ and $h_{\f_1}(1)=1$. The
  mappings $\f_1$, $\f_2$, $\f_4$, $\f_8$ have the same $\Y$-histogram
  and are therefore in the same basis class
  $B_{h_{\f_1}}=\{\f_1,\f_2,\f_4,\f_8\}$. The set
  $\{\f_1,\f_2,\f_4,\f_8,\f_{15}\}$ is \clup\ and corresponds to
  $B_{h_{\f_1}} \cup B_{h_{\f_{15}}}$.
\end{example}

It holds:
\begin{lemma}[\cite{igel:02c}]\label{lemma:lem}
  \begin{enumerate}
  \item\label{a:lem}  Any subset $F\subseteq \F$ that is c.u.p.\ is uniquely defined by
    a union of pairwise disjoint basis classes.
  \item\label{b:lem} $B_h$ is equal to the permutation orbit of any function $f$
    with histogram $h$, i.e.,
    \begin{equation}
      B_h = \bigcup_{\pi\in\pot(\X)} \{ f \circ \pi \} \enspace.
      \end{equation}
\end{enumerate}
\end{lemma}
A proof is  given in \cite{igel:02c}.

\section{Time to Find a Desirable Solution}\label{timetodesire:sec}
Theorem \ref{nfl:thm} tells us that on average all algorithms need the 
same time to find a desirable, say optimal,
solution---but how long does it take?
The average number of evaluations, i.e., the mean first hitting time
$\text{E}\{{\mathcal T}\}$, needed to find an optimum depends on the
cardinality of the search space $|\geno|$ and the number $\opt$ of
search points that are mapped to a desirable solution.

Let $F_{\opt}\subset \F$ be the set of all functions where $\opt$
elements in $\geno$ are mapped to optimal solutions.  For
\emph{non-repeating} black-box search algorithms it holds:

\begin{theorem}[\cite{igel:02f}]
  Given a search space  of cardinality $\x$ the expected
  number of evaluations $\text{E}\{\ti\}$ averaged over
  $F_{\opt}\subseteq\F$ is given by
 \begin{equation}
 \text{E}\{\ti\} =\frac{|\geno| + 1}{\opt + 1}\enspace.
 \end{equation}
 \end{theorem}
 A proof can be found in \cite{igel:02f}, where this result is used to
 study the influence of neutrality (i.e., of non-injective
 genotype-phenotype mappings) on the time to find a desirable solution.

\section{Fraction of Subsets Closed under Permutation}
The NFL-theorems can be regarded as the basic skeleton of
combinatorial optimization and are important for deriving theoretical
results as the one presented in the previous section. However, are the
preconditions of the NFL-theorems ever fulfilled in practice? How likely is it that 
a randomly chosen subset is \clup?

There exist $2^{\left(|\Y|^{|\X|}\right)}-1$ non-empty subsets of
$\F$ and it holds:
\begin{theorem}[\cite{igel:02c}]\label{main:thm}
The number of non-empty subsets of $\Y^\X$ that are \clup{}
is given by
\begin{equation}
2^{\binom{|\X|+|\Y|-1}{|\X|}} -1
\end{equation}
and therefore the fraction of non-empty
 subsets \clup{} is given by
\begin{equation}
\left({2^{\binom{|\X|+|\Y|-1}{|\X|}} -1}\right)\Big/\left({2^{\left(|\Y|^{|\X|}\right)}-1}\right)
\enspace.
\end{equation}
\end{theorem}
\noindent The proof is given in  \cite{igel:02c}.

\begin{figure}[ht!]
\small
\psfrag{10}[l][l]{}
\psfrag{A}[l][l]{$|\Y|=2$}
\psfrag{B}[l][l]{$|\Y|=3$}
\psfrag{C}[r][r]{$|\Y|=4$}
\psfrag{-5}[r][r]{$10^{-5}$}
\psfrag{-10}[r][r]{$10^{-10}$}
\psfrag{-15}[r][r]{$10^{-15}$}
\psfrag{-20}[r][r]{$10^{-20}$}
\psfrag{-25}[r][r]{$10^{-25}$}
\psfrag{ylabel}[][]{\!\!\!\!fraction of subsets \clup{}\rule[-8ex]{0pt}{0pt}}
\psfrag{-30}[r][r]{$10^{-30}$}
\psfrag{-35}[r][r]{$10^{-35}$}
\psfrag{-40}[r][r]{$10^{-40}$}
\psfrag{-45}[r][r]{$10^{-45}$}
\psfrag{-50}[r][r]{$10^{-50}$}
\psfrag{0}[][r]{1}
\psfrag{1}[c][c]{1}
\psfrag{2}[c][c]{2}
\psfrag{3}[c][c]{3}
\psfrag{4}[ct][ct]{$\underset{\displaystyle |\X|}{4}$}
\psfrag{5}[c][c]{5}
\psfrag{6}[c][c]{6}
\psfrag{7}[c][c]{7}
\psfrag{8}[c][c]{8}
\begin{center}
\epsfig{file=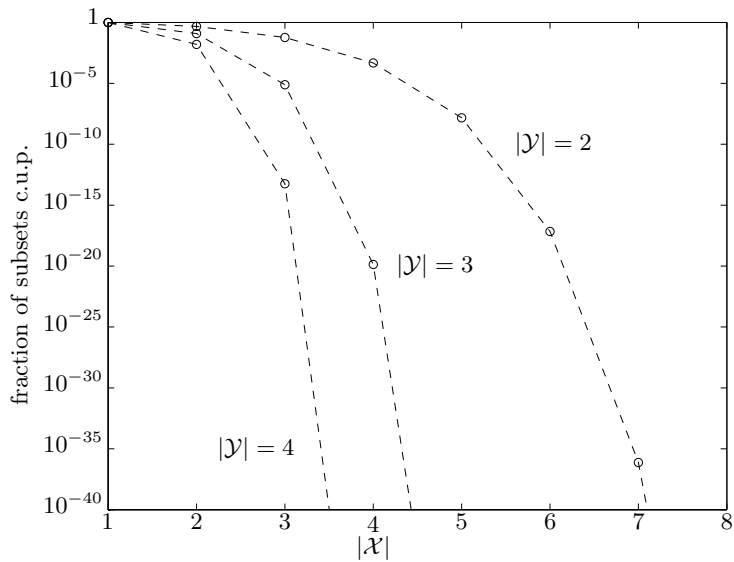,width =.75\textwidth}
\end{center}\smallskip

\caption{\label{plot:fig}
 The ordinate gives the fraction of subsets closed under permutation
 on logarithmic scale
 given the cardinality of the search space $\X$.
 The different curves correspond to different cardinalities
 of the codomain $\Y$.}
\end{figure}

Figure \ref{plot:fig} shows a plot of the fraction of non-empty subsets \clup\ 
versus the cardinality of $\X$ for different values of $|\Y|$.  The fraction
decreases for increasing $|\X|$ as well as for increasing $|\Y|$.  More
precisely, for $|\Y|> e|\X|/(|\X| - e)$ it converges to zero double
exponentially fast with increasing $|\X|$.  Already for small $|\X|$ and
$|\Y|$ the fraction almost vanishes.

Thus, the statement ``I'm only interested in a subset $F$ of all
possible functions, so the precondition of the sharpened NFL-theorems
is not fulfilled'' is true with a probability close to one (if $F$ is
chosen uniformly and $\Y$ and $\X$ have reasonable cardinalities).
The fact that the precondition of the NFL-theorem is violated does not
lead to ``Free Lunch'', but nevertheless ensures the possibility
of a ``Free Appetizer''.

\section{Search Spaces with Neighborhood Relations}\label{toppology:sec}
Although the fraction of subsets \clup\ is close to zero already for
small search and cost-value spaces, the absolute number of subsets
\clup\ grows rapidly with increasing $|\X|$ and $|\Y|$.  What if these
classes of functions are the relevant ones, i.e., those we are
dealing with in practice?  

Two assumptions can be made for most of the functions relevant in
real-world optimization: First, the search space has some structure.
Second, the set of objective functions fulfills some constraints
defined based on this structure.  More formally, there exists a
non-trivial neighborhood relation on $\X$ based on which constraints
on the set of functions under consideration are formulated, e.g., concepts
like ruggedness or local optimality and constraints like upper bounds
on the ruggedness or on the maximum number of local minima can be
defined.  

A neighborhood relation on $\X$ is a symmetric
function $n:\X\times\X\to\{0,1\}$.  Two elements $x_i,x_j\in\X$ are
called neighbors iff $n(x_i,x_j)=1$.  A neighborhood 
relation is called
non-trivial iff $\exists x_i,x_j\in\X: x_i\neq x_j \,\wedge\,
n(x_i,x_j)=1$ and $\exists x_k,x_l\in\X: x_k\neq x_l \,\wedge\,
n(x_k,x_l)=0$.  It holds:
\begin{theorem}[\cite{igel:02c}]
  A non-trivial neighborhood relation on $\X$ is not invariant under
  permutations of $\X$.
\end{theorem}

This result is quite general.  Assume that the search space $\X$ can
be decomposed as $\X=\X_1 \times \dots\times \X_l, l>1$, and let on one
component $\X_i$ exist a non-trivial neighborhood
$n_i:\X_i\times\X_i\to\{0,1\}$. This neighborhood induces a
non-trivial neighborhood on $\X$, where two points are neighbored iff
their $i$-th components are neighbored with respect to $n_i$. Thus,
the constraints discussed below need only refer to a single component.
Note that the neighborhood relation need not be the canonical one
(e.g., Ham\-ming-distance for Boolean search spaces).  For example, if
integers are encoded by bit-strings, then the bit-strings can be
defined as neighbored iff the corresponding integers are.

Some constraints that are defined with respect to a neighborhood
relation and that are relevant in practice are now discussed,
cf.~\cite{igel:02c}. For this purpose, a metric
$\m_{\Y}:\Y\times\Y\rightarrow\dR$ on $\Y$ is presumed, e.g., in the
typical case of real-valued target functions $\Y\subset\dR$ the
Euclidean distance.

A constraint on steepness leads to a set of functions that is not
\clup{} Based on a neighborhood relation on the search space, we can
define a simple measure of maximum steepness of a function $f\in\F$ by
the maximum distance of the target values of neighbored points
$s^{\max}(f)=\max_{x_i,x_j\in\X\,\wedge\, n(x_i,x_j) = 1 }
\m_{\Y}(f(x_i),f(x_j))$.  Further, for a function $f\in F$, the
diameter of its range can be defined as
$d^{\max}(f)=\max_{x_i,x_j\in\X} \m_{\Y}(f(x_i),f(x_j))$.
\begin{corollary}[\cite{igel:02c}]
  If the maximum steepness $s^{\max}(f)$ of every function $f$ in a
  non-empty subset $F\subset \F$ is constrained to be smaller than the
  maximal possible $\max_{f\in F} d^{\max}(f)$, then $F$ is not \clup
\end{corollary}

Consider the number of local minima, which
is often regarded as a measure of complexity \cite{whitley:99}.  For a
function $f \in \F$ a point $x \in \X$ is a local minimum iff
$f(x)<f(x_i)$ for all neighbors $x_i$ of $x$.  Given a function $f$
and a neighborhood relation on $\X$, let $l^{\max}(f)$ be the
maximal number of minima that functions with the same $\Y$-histogram
as $f$ can have (i.e., functions where the number of $\X$-values that
are mapped to a certain $\Y$-value are the same as for $f$). 
\begin{corollary}[\cite{igel:02c}]
  If the number of local minima of every function $f$ in a non-empty
  subset $F\subset \F$ is constrained to be smaller than the maximal
  possible $\max_{f\in F} l^{\max}(f)$, then $F$ is not \clup
\end{corollary}

\begin{example} Consider all mappings $\{0,1\}^\ell\to\{0,1\}$ that
  have less than the maximum number of $2^{n-1}$ local minima w.r.t.\ 
  the ordinary hypercube topology on $\{0,1\}^\ell$. This means, this
  set does not contain mappings such as the parity function, which
  is one iff the number of ones in the input bitstring is even. This
  set is not \clup
\end{example}

Hence, statements like ``In my application domain, functions with
maximum number of local minima are not realistic'' and ``For some
components, the objective functions under consideration will not have
the maximal possible steepness'' lead to scenarios where the
precondition of the NFL-theorem is not fulfilled.

\section{A Non-Uniform NFL-theorem}\label{nonuni:sec}

In the sharpened NFL-theorem it is implicitly presumed
that all functions in the subset $F$ are equally likely since
averaging is done by uniform summation over $F$. Here, we investigate
the general case when every function $f \in \F$ has an arbitrary
probability $p(f)$ to be the objective function. Such a
non-uniform distribution of the functions in $F$ appears to be much
more realistic. Until now, there exist only very weak results for this
general scenario. For example, let for all $x\in\X$ and $y\in\Y$
\begin{equation}
p_{x}(y):=  \sum_{f\in\F} p(f)\,\delta(f(x),y)\enspace,
\end{equation}
i.e., $p_{x}(y)$ denotes the probability that the search point $x$
is mapped to the cost-value $y$.  In \cite{english:00} it has been
shown that a NFL-result holds if within a class of functions the
function values are i.i.d.{}, i.e., if
\begin{equation}\label{english1:eq}
\forall x_1,x_2\in\X: p_{x_1}=p_{x_2}\text{ and } p_{x_1, x_2}=p_{x_1} p_{x_2}\enspace,
\end{equation}
where $p_{x_1, x_2}$ is the joint probability distribution of the
function values of the search points $x_1$ and $x_2$. However, this is
not a necessary condition and applies only to extremely
``unstructured'' problem classes.

The following theorem gives a   necessary and sufficient condition for
a NFL-result in the general case of non-uniform distributions:
\begin{theorem}[non-uniform sharpened NFL]\label{nonuninfl:thm}
For any two  algorithms $a$ and $b$,
any value $k\in \dR$, and any performance measure $c$
\begin{equation}\label{nonu:eq}
\sum_{f\in \F}p(f)\,\delta(k, c(Y(f,m,a)))
=
\sum_{f\in \F}p(f)\,\delta(k, c(Y(f,m,b)))
\end{equation}
iff for all $h$
\begin{equation}\label{nonu2:eq}
f,g\in B_{h}\Rightarrow p(f)=p(g)\enspace.
\end{equation}
\end{theorem}
\begin{proof}
First, we show that (\ref{nonu2:eq}) implies that (\ref{nonu:eq})
holds for any $a$, $b$, $k$, and $c$.
It holds by lemma \ref{lemma:lem}(\ref{a:lem})
\begin{align}
\sum_{f\in \F}p(f)\,\delta(k, c(Y(f,m,a)))
& = \sum_{h\in\H} \sum_{f\in B_h}p(f)\,\delta(k, c(Y(f,m,a)))
\intertext{using $f,g\in B_{h}\Rightarrow p(f)=p(g)=p_h$}
& = \sum_{h\in\H} p_h \sum_{f\in B_h}\delta(k, c(Y(f,m,a)))
\intertext{as each $B_h$ is \clup{} we may use theorem \ref{nfl:thm}}
& = \sum_{h\in\H} p_h \sum_{f\in B_h}\delta(k, c(Y(f,m,b)))\\
& = \sum_{f\in \F}p(f)\,\delta(k, c(Y(f,m,b)))
\enspace.
\end{align}
Now we prove that (\ref{nonu:eq}) being true for any $a$, $b$, $c$,
and $k$ implies (\ref{nonu2:eq}) by showing that if (\ref{nonu2:eq})
is not fulfilled then there exist $a$, $b$, $c$, and $k$ such that
(\ref{nonu:eq}) is also not valid.
Let $f,g\in B_{h}$, $f\neq g$, $p(f)\neq p(g)$, and $g=f\circ\pi$.
Let $\X=\{{\xi}_1,\dots,{\xi}_n\}$.  Let $a$ be an algorithm that
always enumerates the search space in the order
${\xi}_1,\dots,{\xi}_n$ regardless of the observed cost-values and let
$b$ be an algorithm that enumerates the search space always in the
order $\pi^{-1}({\xi}_1),\dots,\pi^{-1}({\xi}_n)$.  It holds
$g(\pi^{-1}(\xi_i))=f(\xi_i)$ for $i=1,\dots,n$ and
$Y(f,n,a)=Y(g,n,b)$.  We consider the performance measure
\begin{equation}
c^\dagger(\left<y_1,\dots,y_m\right>)=\begin{cases}1 & \text{if } m=n \wedge
\left<y_1,\dots,y_m\right>=\left<f({\xi}_1),\dots,f({\xi}_n)\right>\\
0 & \text{otherwise}
\end{cases}
\end{equation}
for any $y_1,\dots,y_m\in\Y$. Then, for $m=n$ and $k=1$, we have
\begin{equation}
\sum_{f'\in \F}p(f')\,\delta(k, c^\dagger(Y(f',n,a))) = p(f)
\enspace,
\end{equation}
as $f'=f$ is the only function $f'\in \F$ that yields
\begin{equation}
\left<f'({\xi}_1),\dots,f'({\xi}_n)\right> =
 \left<f({\xi}_1),\dots,f({\xi}_n)\right>\enspace,
\end{equation}
and
\begin{equation}
\sum_{f'\in \F}p(f')\,\delta(k, c^\dagger(Y(f',n,b))) = p(g)
\enspace,
\end{equation}
and therefore (\ref{nonu:eq}) does not hold.
\qed
\end{proof}

The sufficient condition given in \cite{english:00} is a special case
of theorem \ref{nonuninfl:thm}, because (\ref{english1:eq}) implies
\begin{equation}
g=f\circ\pi\Rightarrow p(f) = p(g)
\end{equation}
for any $f,g\in\F$ and $\pi\in\pot(\X)$, which in turn implies $g,
f\in B_{h} \Rightarrow p(f) = p(g)$ due to lemma
\ref{lemma:lem}(\ref{b:lem}).

The probability that a randomly chosen distribution over the set of
objective functions fulfills the preconditions of theorem
\ref{nonuninfl:thm} has measure zero.  This means that in this general
and realistic scenario the probability that the conditions for a
NFL-result hold vanishes.

\section{Conclusion}

Several recent results on NFL-theorems for optimization presented in
\cite{igel:02c,igel:02f} were summarized and extended.  In particular,
we derived necessary and sufficient conditions for NFL-results for
arbitrary distributions of target functions and thereby presented the
``sharpest'' NFL theorem so far.  It turns out that in this
generalized scenario, the necessary conditions for NFL-results can not
be expected to be fulfilled.

\subsection*{Acknowledgments}

This work was supported by the DFG, grant Solesys-II SCHO 336/5-2.  We
thank Stefan Wiegand for fruitful discussions.

{
\small
\bibliographystyle{abbrv} \bibliography{igelToussaint} }

\end{document}